\def\paperlanguage{}
\pdfoutput=1 

\documentclass[letterpaper, 10 pt, conference]{ieeeconf}  

\usepackage{bm}
\usepackage{cite}
\usepackage{flushend}
\include{preamble}

\IEEEoverridecommandlockouts                              

\title{\LARGE \textbf
  {
    \switchlanguage%
    {%
      Human Mimetic Forearm Design with Radioulnar Joint\\
      using Miniature Bone-Muscle Modules and Its Applications
    }%
    {%
      骨構造一体小型筋モジュールにより構成された\\
      人体解剖学に基づく橈骨尺骨構造を有する前腕の開発
    }%
  }
}

\author{Kento Kawaharazuka, Shogo Makino, Masaya Kawamura, \\Yuki Asano, Yohei Kakiuchi, Kei Okada and Masayuki Inaba
  \thanks{Authors are with Department of Mechano-Informatics, Graduate School of Information Science and Technology, The University of Tokyo, 7-3-1 Hongo, Bunkyo-ku, Tokyo, 113-8656, Japan.
    {\texttt\small [kawaharazuka, makino, kawamura, asano, youhei, k-okada, inaba]@jsk.t.u-tokyo.ac.jp}
  }
}

\begin{document}

\maketitle
\thispagestyle{empty}
\pagestyle{empty}

\begin{abstract}
  \switchlanguage%
  {%
    The human forearm is composed of two long, thin bones called the radius and the ulna, and rotates using two axle joints.
    We aimed to develop a forearm based on the body proportion, weight ratio, muscle arrangement, and joint performance of the human body in order to bring out its benefits.
    For this, we need to miniaturize the muscle modules.
    To approach this task, we arranged two muscle motors inside one muscle module, and used the space effectively by utilizing common parts.
    In addition, we enabled the muscle module to also be used as the bone structure.
    Moreover, we used miniature motors and developed a way to dissipate the motor heat to the bone structure.
    Through these approaches, we succeeded in developing a forearm with a radioulnar joint based on the body proportion, weight ratio, muscle arrangement, and joint performance of the human body, while keeping maintainability and reliability.
    Also, we performed some motions such as soldering, opening a book, turning a screw, and badminton swinging using the benefits of the radioulnar structure, which have not been discussed before, and verified that Kengoro can realize skillful motions using the radioulnar joint like a human.
  }%
  {%
    人間の前腕は橈骨と尺骨という二つの細長い骨が車軸関節を通して回転した構造となっている。
    我々はこのメカニズムの利点を引き出すため、人体プロポーションや重量、関節トルクや可動域を重視した前腕の設計を目指した。
    そのためには筋モジュールの小型化が必要である。
    そのためのアプローチとして、1モジュールに2つの筋を配置し、部品の共通化などによって空間を有利に使い、筋を骨格としても使えるようにした。
    また、小型モータを用い、その出力的ビハインドを補うために骨格に熱を逃がす構造を考えた。
    骨構造一体小型筋モジュールというアプローチによって今まで実現していなかった、メンテナンス性と信頼性を保ちつつ、
    多自由度であり、人体のプロポーションや重量、関節性能を模倣した橈骨尺骨構造を有する前腕の開発に成功した。
    また、今まで議論されてこなかった橈尺関節の良さを活かしたハンダ付け動作、本を開く動作、ねじ回し動作、スイング動作を行い、
    人間と同様な、橈骨尺骨構造を巧みに使った動作が実現できることを確認した。
  }%
\end{abstract}

\section{INTRODUCTION} \label{sec:1}
\switchlanguage%
{%
  In recent years, development of the humanoid is vigorous.
  The humanoid, beginning with the ASIMO \cite{icra1998:hirai:asimo}, has two arms and two legs, and can move and walk like a human.
  The development of not only the humanoid, but of the tendon-driven musculoskeletal humanoid, which is based on various parts of the human body, is also vigorous \cite{humanoids2010:hugo:eccerobot, humanoids2016:asano:kengoro}.
  The tendon-driven musculoskeletal humanoid is based on not only the body proportion but also the joint structure, drive system, and muscle arrangement of the human body, and is used to analyze human motion and to achieve human skillful motion.
  Of these studies, there are many which duplicate the human joint structure.
  Asano, et al. duplicates the human screw home mechanism, and discusses the achievement of motion using this structure \cite{iros2012:asano:knee}.
  Also, Sodeyama, et al. discusses the design of the upper limb using the clavicle and scapula \cite{iros2007:sodeyama:kojiro-shoulder}.
  Like so, there are many studies that integrate structures specific to humans with humanoids.
}%
{%
  近年、人型をしたヒューマノイドの開発が盛んに行われている。
  ASIMO\cite{icra1998:hirai:asimo}を始めとしたヒューマノイドは二腕二脚を有し、人間らしい動作、歩行が可能である。
  また、人型というだけでなく、
  人体のあらゆる部分を模倣した筋骨格腱駆動ヒューマノイドの開発も盛んになっている\cite{humanoids2010:hugo:eccerobot, humanoids2016:asano:kengoro}。
  筋骨格ヒューマノイドにおいては、プロポーションだけでなく、関節構造や関節駆動方法、筋配置や重量までも模倣し、人間の動作分析や、人間の巧みな動作実現を試みている。
  その中でも、人体の関節構造を模した研究は多い。
  浅野らは人体の膝の終末回旋機構を模し、その動作実現性について論じている\cite{iros2012:asano:knee}。
  また、袖山らは鎖骨と肩甲骨による上肢の設計について論じている\cite{iros2007:sodeyama:kojiro-shoulder}。
  このように、人間特有の構造をヒューマノイドに組み込むという研究は多く存在する。
}%
\begin{figure}[htb]
  \centering
  \includegraphics[width=1.0\columnwidth]{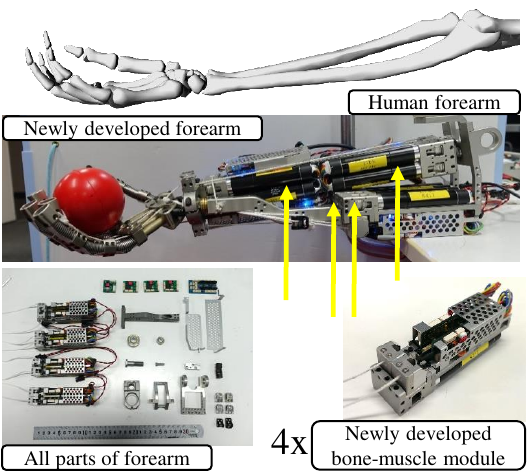}
  \vspace{-3.0ex}
  \caption{Forearm of Kengoro, composed of newly developed miniature bone-muscle module.}
  \label{figure:kengoro-and-radioulnar}
  \vspace{-2.0ex}
\end{figure}
\begin{figure*}[htb]
  \centering
  \includegraphics[width=1.85\columnwidth]{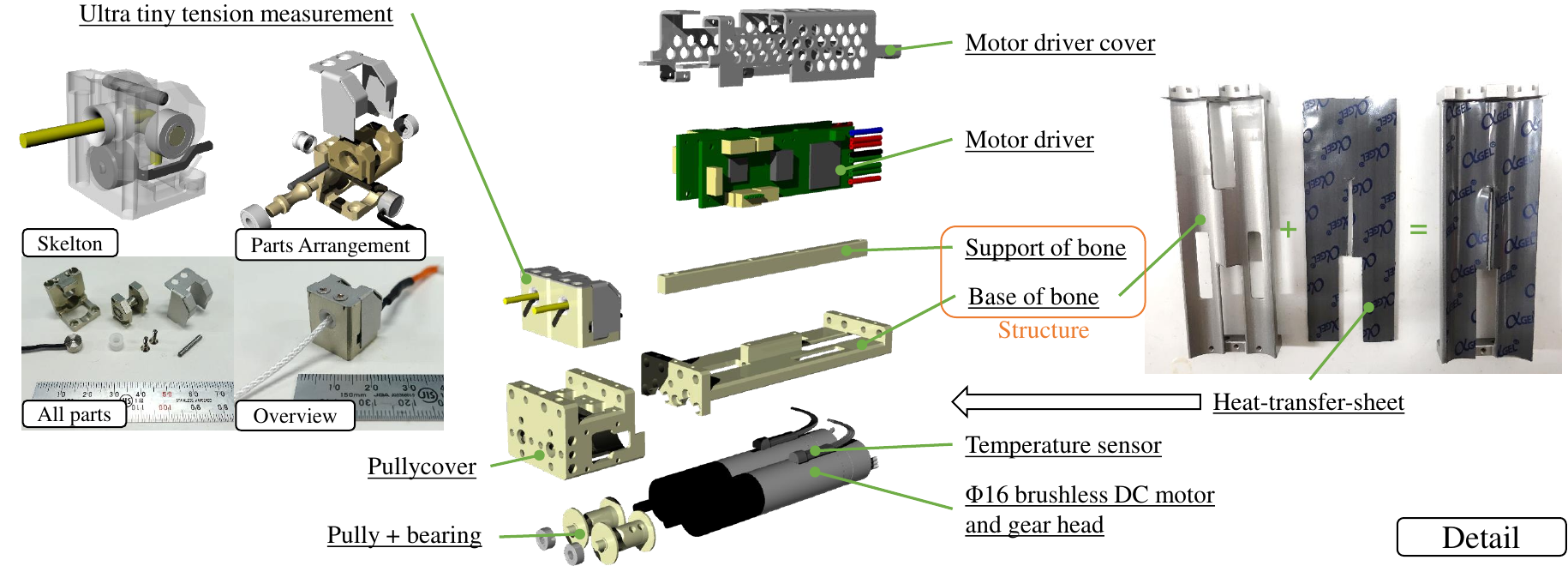}
  \caption{Details of the newly developed miniature bone-muscle module.}
  \label{figure:double-motor-module-detail}
  \vspace{-2.0ex}
\end{figure*}
\switchlanguage%
{%
  On the other hand, there are few studies which discuss the human specific radioulnar joint structure.
  Some examples of humanoids with a radioulnar joint are \cite{iros2012:ikemoto:shoulder, robio2011:nakanishi:kenzoh}, but these are made of pneumatic actuators that are easy to arrange but have poor controllability, or are unable to arrange the number of muscles needed to achieve many DOFs in the forearm.
  The conventional method of installing the muscle modules such as \cite{iros2015:asano:module, humanoids2013:michael:anthrob} to the structure excels in maintainability and reliability, and includes electric motors, which have better controllability.
  However, we need to miniaturize the modules or propose other approaches in order to achieve many DOFs without deviating from the human body proportion, because the conventional muscle modules are large in size, and need other wasteful structures to function.
  Additionally, the muscle arrangements, the proportion of the forearm, and the benefits of the radioulnar structure are not discussed at all in previous studies.

  Thus, in this study, we conduct research about the development of a forearm with a radioulnar joint based on the proportion, weight ratio, muscle arrangement, joint structure, and joint performance of the human body, and about the motions that use its structure skillfully.
  Then, we developed a new miniature bone-muscle module, which integrates a muscle module with the structure.
  By using this miniature bone-muscle module, we can achieve the human mimetic forearm with a radioulnar joint while keeping many DOFs, maintainability, and reliability.
  Then, we succeeded in achieving human skillful motion, which makes the best use of the radioulnar structure, but has not been discussed before.

  In \secref{sec:1}, we explained the motive and goal of this study.
  In \secref{sec:2}, we will explain the development and performance of the miniature bone-muscle module necessary for the forearm with a radioulnar joint.
  In \secref{sec:3}, we will explain the achievement of the radioulnar structure using miniature bone-muscle modules, and evaluate the degree of imitation.
  In \secref{sec:4}, we will discuss experiments of soldering, opening a book, turning a screw, and badminton swinging as examples of human skillful motion that use the benefits of the radioulnar joint.
  Finally in \secref{sec:5}, we will state the conclusion and future works.
}%
{%
  しかし、人間特有の構造である橈骨尺骨構造について論じた研究は少ない。
  橈骨尺骨構造のついたヒューマノイドとしては\cite{iros2012:ikemoto:shoulder, robio2011:nakanishi:kenzoh}が挙げられるが、
  制御性は悪いが配置の容易な空気圧アクチュエータによるものや、前腕に筋を多く配置できずに多自由度を実現できていないものばかりである。
  モータで駆動され、メンテナンス性や信頼性に優れた~や~の筋モジュールによって前腕部を構成する方法が挙げられが、プロポーションを崩さずに多自由度を実現するためには筋モジュールの小型化やその他のアプローチが必要である。
  また、筋配置やプロポーション、その構造自体の利点に関する議論は全く行われていない。

  そこで本研究では, 人体のプロポーションや重量, 筋配置や構造などを正しく模倣した橈骨尺骨構造を有する前腕部の設計開発と, それを巧みに用いた動作に関する研究を行う.
  設計開発のアプローチとして, 我々は腱駆動筋モジュールと骨格を融合させた, 骨構造一体小型筋モジュールを開発した.
  この骨構造一体小型筋モジュールを用いることで, メンテナンス性を保ちつつ人体模倣型橈骨尺骨構造を人体のプロポーションや重量, 筋配置を模倣しつつ実現することができた.
  それによって我々は, 今まで議論されて来なかった橈骨尺骨構造の利点を活かした巧みな動作を実現することにも成功した.

  \secref{sec:1}では本論文の動機と目標について述べた。
  \secref{sec:2}では人体模倣型橈骨尺骨構造を有する前腕のために必要な骨構造一体小型筋モジュールの開発、性能について述べる。
  \secref{sec:3}ではこの筋モジュールを用いた橈骨尺骨構造の実現と模倣度評価を行う。
  \secref{sec:4}ではこの前腕の利点を用いた人体の巧みな動作例として、ハンダ付け動作、本開け動作、ネジ回し動作、スイング動作を行う。
  最後に、\secref{sec:5}では結論と今後の方針について述べる。
}%

\section{Development of Miniature Bone-Muscle Module} \label{sec:2}
\switchlanguage%
{%
  \subsection{Approach to Miniature Bone-muscle Module}
  The human forearm is composed of two long, thin bones.
  These bones are called the radius and the ulna, and the radioulnar structure is composed of these two bones and two axle joints located at the proximal and distal.
  However, the actualization of the radioulnar structure is not easy.
  We have developed tendon-driven musculoskeletal humanoids such as Kojiro \cite{humanoids2007:mizuuchi:kojiro}, Kenzoh \cite{robio2011:nakanishi:kenzoh}, and Kenshiro \cite{humanoids2012:nakanishi::kenshirodesign}, but these were unable to completely realize the radioulnar joint, radiocarpal joint and interphalangeal joints.
  This is due to the arrangement of muscles.
  Conventionally, the body of the tendon-driven musculoskeletal humanoid is made by installing muscle modules with actuators, sensors, and circuits to the bone structure.
  For example, there are muscle modules such as Kengoro's module\cite{iros2015:asano:module} and Anthrob's module\cite{humanoids2013:michael:anthrob}.
  This method of installing muscle modules is very effective from the viewpoint of maintainability, reliability, and versatility.
  However, because the radioulnar structure is composed of two long, thin bones, if we install muscle modules to the bone structure, the forearm will be out of proportion, and it will be very difficult to imitate the human body in detail using many muscle modules.
  Thus, we developed a new miniature bone-muscle module.
  We succeeded in developing this muscle module using the two strategies shown below.
  \begin{itemize}
    \item Integration of Muscle and Bone

      This muscle module includes two actuators.
      This approach creates space among the two motors, and we are able to make use of this space.
      Also, the benefit of utilizing common parts for the two muscles is big in saving space.
      We can arrange parts of the bone structure in this space.
      Thus, this muscle module integrates muscle actuators to the bone structure, allowing compact arrangement without wastefully separating the structure from the muscle modules.

    \item Adoption of Miniature Motors and Heat Dissipation by Adherence between the Muscle and Structure

      It is easiest to use small motors as muscles in order to make muscle modules compact.
      However, it is not a good idea to equip a high gear ratio motor for high torque, considering the backdrivability and efficiency.
      Additionally, miniature muscle motors heat up easily.
      To compensate for such drawbacks of adopting miniature motors, this module can keep continuous high tension by dissipating the muscle heat to the structure through a heat transfer sheet.

  \end{itemize}
  Through these approaches, we propose that we can actualize the radioulnar structure based on the body proportion, weight ratio, and muscle arrangement of the human body by simply connecting the muscle modules linearly, which can act as not only the muscle but also as the structure.
  In related works, for an ordinary robot, the integration of frameless motors into the structure is being developed as adopted in TORO \cite{humanoids2016:johannes:toro}.
  Additionally, we aim to develop high maintainability and reliability of the module by packaging motor drivers, sensors, and cables, like the sensor-driver integrated muscle module \cite{iros2015:asano:module}.
  At the same time, by preparing versatility in the arrangement of muscle modules, we propose that we can use this module for not only radioulnar joints, but also for all next-generation tendon-driven musculoskeletal humanoids.
}%
{%
  \subsection{骨構造一体小型筋モジュールのアプローチ}
  人間の前腕は二本の細い骨で構成されている。 これらは橈骨・尺骨と呼ばれ、これらの近位と遠位にある車軸関節によって橈尺関節を成している。
  橈骨尺骨構造を実現するのはそう容易ではない。
  我々の研究室では\cite{robio2011:nakanishi:kenzoh, humanoids2007:mizuuchi:kojiro, humanoids2012:nakanishi::kenshirodesign}という筋骨格腱駆動ヒューマノイドを
  開発してきたが、そのどれも人体の橈尺関節、橈骨手根関節、指関節をまともに実現できた前腕は存在しない。
  それは、筋の配置に問題がある。
  従来の筋モジュールは、アクチュエータや基板を含んだモジュールを骨格に取り付けることで人体を模してきた。
  例として\cite{iros2015:asano:module}や\cite{humanoids2013:michael:anthrob}のような筋モジュールが挙げられ、
  これらは骨格となる部品にこの筋モジュールを取り付けることで体を成す。
  しかし、橈骨尺骨は二本の細い骨で構成されており、この骨格に筋モジュールを取り付けてしまっては橈骨尺骨のプロポーションを崩してしまううえ、
  多数の筋モジュールによってより詳細な人体模倣を行うことは困難である。
  そこで我々は、新しい骨構造一体小型筋モジュールを開発した。
  この筋モジュールは以下の二つの戦略を取ることでよりコンパクトな筋モジュールの開発に成功している。
  \begin{itemize}
    \item 筋と骨格の一体化

      筋モジュール一つに対して二つの筋を格納している。これによってモータとモータの間に空間ができ、それを活用することができる。
      また、部品を共通化することによる体積的メリットは大きい。
      その空いた空間に骨格としての部品を配置している。
      これによって、骨格と筋が一体となり、無駄に骨格と筋モジュールを分けず、よりコンパクトに全体を収めることが可能となる。

    \item モジュール自体の小型化

      モジュールをコンパクトにするためには、筋となるモータを小さいものにするのが最も手っ取り早い。
      しかし、大きなトルク得るために大きなギア比のモータを取り付けるのはバックドライバビリティや効率の点から良いとは言えない。
      また、小型化によってモータがすぐに熱を帯びてしまう。
      そのような小型モータの採用によるビハインドを補うため、骨格に伝熱性のシートを通して筋の熱を逃がすことで持続的高張力を確保する。
  \end{itemize}
  これら二つのアプローチにより、骨格にもなる筋モジュールを単純に縦に骨として繋げることによって橈骨尺骨構造を実現することができ、
  人体プロポーション、筋配置、重量を模倣することができると考える。
  通常の軸駆動型ロボットではTORO\cite{humanoids2016:johannes:toro}で採用されているようなフレームレスモータを骨格に埋め込む方式が開発されている。
  また、\cite{iros2015:asano:module}の筋モジュールのようにMotor Driverやケーブルをパッケージ化することにより、非常に信頼性、メンテナンス性の高い筋モジュールを目指した。
  同時に、様々な配置が可能な汎用性を備えることによって、橈骨尺骨構造だけでなく、次世代の筋骨格腱駆動ロボットすべてに応用できると考える。
}%


\begin{table*}[t]
  \centering
  \caption{Comparison of newly developed miniature bone-muscle module and sensor driver integrated muscle module \cite{iros2015:asano:module}.}
  \scalebox{0.9}{
    \begin{tabular}{c|c|c}
      \hline
      & Miniature bone-muscle module in this study & Sensor-driver integrated muscle module \cite{iros2015:asano:module}
      \\ \hline \hline
      Module dimension [${\textrm{mm}}^{3}$] & $32.0\times40.0\times126$ & $22.0\times40.5\times149$ \\ \hline
      Module weight [kgf] & 0.30 & 0.32 \\ \hline
      Number of actuators & 2 & 1 \\ \hline
      Actuator & BLDC-60W (changeable) & BLDC-120W (changeable) \\ \hline
      Diameter of winding pulley [mm] & 8 & 12 \\ \hline
      Reduction ratio of actuator & 157:1 (changeable) & 53:1 (changeable) \\ \hline
      Continuous maximum winding tension [N] & 424 & 338 \\ \hline
      Winding rate with no load [mm/s] & 116 & 200 \\ \hline
    \end{tabular}
  }
  \label{table:comparison-of-muscle-module}
\end{table*}

\switchlanguage%
{%
  \subsection{Development Details of Miniature Bone-muscle Module}

  The details of the miniature bone-muscle module are shown in \figref{figure:double-motor-module-detail}.
  The motor is a brushless DC motor, and we use 84:1 or 157:1 as the gear ratio of the motor depending on the muscle.
  The wire is Dyneema and is winded up by the $\phi8$ pulley.
  The cables from the load cell of the tension measurement unit, temperature sensor attached to the motor, and hall sensor of the motor are all connected to the motor driver, and a cover protects these cables and circuits, increasing operational stability.

  We especially would like to discuss three topics.
  First, ``Support of bone'' and ``Base of bone'' become the bone structure, enabling the use of the muscle module as the structure.
  Thus, we are able to connect the muscle modules lengthwise and crosswise as the structure, eliminating waste.
  Second, this module can dissipate heat to the structure through the heat transfer sheet between ``Base of bone'' and the two motors.
  As a result, the module can realize comparatively continuous high muscle tension even if the motor is miniature and the gear ratio is 84:1 or 157:1, which we can backdrive.
  Finally, we developed an ultra tiny tension measurement unit.
  We can use space effectively by arranging the load cell, which defines the size of the unit, vertically.
  We succeeded in decreasing the volume to 61$\%$ compared to the old tension measurement unit \cite{iros2015:asano:module}.
  The size of this unit is $16\times16\times19$ [${\textrm{mm}}^3$] and is designed to measure tension until 56.5 [kgf].

  \subsection{Evaluating Performance of Miniature Bone-muscle Module}

  First, we compare the size, weight, maximum muscle tension, and so on, between the newly developed miniature bone-muscle module and the conventional muscle module \cite{iros2015:asano:module}.
  The result of the comparison is shown in \tabref{table:comparison-of-muscle-module}.
  Since the module developed in this study has two muscle actuators inside one module, and the size and performance of the motors are different between the two modules, a simple comparison cannot be done.
  However, the module developed in this study was able to double the number of muscle with only a 21$\%$ increase in volume.
}%
{%
  \subsection{骨構造一体小型筋モジュールの設計詳細}
  骨構造一体小型筋モジュールの詳細は\figref{figure:double-motor-module-detail}のようになっている。
  モータはブラシレスDCモータを使用し、ギア比は場所によって84:1と157:1を使い分けている。
  ワイヤはDyneemaを使用しており、それをPulleyによって巻き取る方式となっている。
  張力測定ユニットのロードセルやモータに貼り付けた温度センサ、モータからのホールセンサのケーブルはすべてモータドライバに接続されており、
  その配線やMotorDriverを守るためのカバーが接続され動作安定性を向上させている。

  特筆すべきは、3ヶ所である。
  まず、Support of boneとBase of boneが骨格となり、筋モジュール自体を骨としても使えるようになっている。
  つまり、骨格なしでこの筋モジュールを縦や横に並べて骨格として配置することができるのである。
  次に、Base of boneとMotorの間に伝熱性のシートを貼ることによって骨格に熱を逃がすようになっている。
  これにより84:1, 157:1という人間がバックドライブできるギリギリの小型モータでも、比較的持続的高張力を実現する。
  最後に、超小型張力測定ユニットである。
  これは、張力測定ユニットの大きさを規定してしまうロードセルをあえて縦に配置することで空間を有利に使い、今までの張力測定ユニット\cite{iros2015:asano:module}よりも
  体積を61$\%$に減らすことに成功している。サイズは16x16x19[${\textrm{mm}}^3$]であり、55[kgf]までの張力を測定することができるように設計してある。
  完成した骨構造一体小型筋モジュールは\figref{figure:double-motor-module-overview}のようになっている。

  \subsection{骨構造一体小型筋モジュールの性能評価}

  まず、骨構造一体小型筋モジュールと従来の筋モジュール\cite{iros2015:asano:module}で大きさや重量、最大張力などの比較を行う。
  比較結果は\tabref{table:comparison-of-muscle-module}のようになっている。
  本研究で開発した骨構造一体小型筋モジュールには1つのモジュールに2つの筋が含まれており、また、アクチュエータのサイズも性能も異なるため単純な比較はできないが、
  体積の$21\%$の上昇に対して、筋数を二倍にすることができている。
}%

\begin{figure}[htb]
  \centering
  \includegraphics[width=1.0\columnwidth]{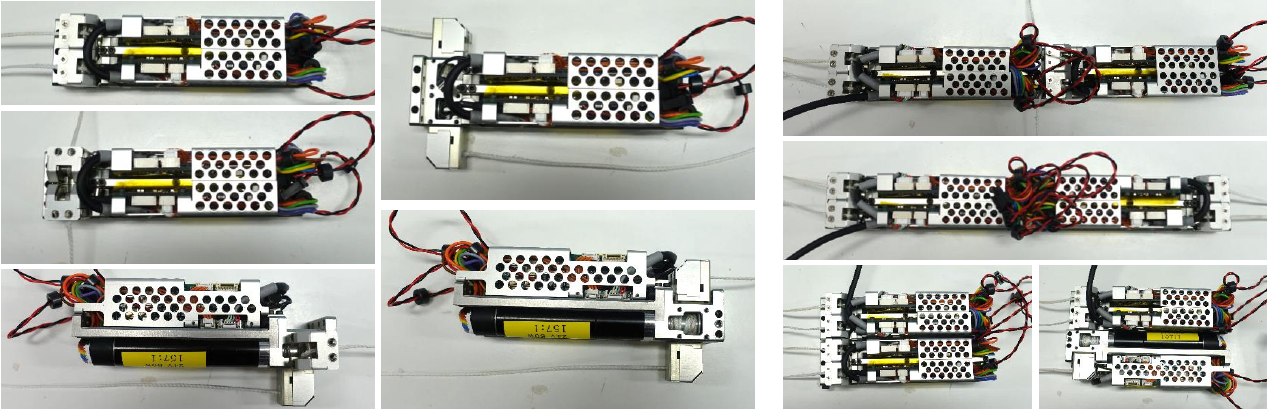}
  \vspace{-3.0ex}
  \caption{General versatility of the newly developed bone-muscle module. Left: various arrangements of ultra tiny tension measurement unit. Right: various connections of muscle modules.}
  \label{figure:various-module-pose}
  \vspace{-2.0ex}
\end{figure}

\switchlanguage%
{%
  Second, we discuss the versatility of the miniature bone-muscle module.
  A characteristic of this muscle module lies in the integration of the muscle and structure, but we must not lose freedom of design of the robot through modularization.
  Thus, this muscle module is designed in a way that makes it possible for the ultra tiny tension measurement units to be arranged in various directions and positions, as shown in the left of \figref{figure:various-module-pose}, to gain freedom in muscle arrangement.
  The connection among modules can also be arranged in various ways as shown in the right of \figref{figure:various-module-pose}, and we can create various designs using the muscle module as the structure.
}%
{%
  次に、骨構造一体小型筋モジュールの汎用性について議論する。
  本研究の特徴は骨構造一体という部分にあるが、筋モジュールの前提要件として、設計の自由度を殺さないということが挙げられる。
  そのため、骨構造一体小型筋モジュールに張力測定ユニットは\figref{figure:various-module-pose}のように様々方向に取り付けることができるように設計されており、
  筋配置の自由度を得ている。
  同様に、筋モジュール間の結合も\figref{figure:various-module-pose}のように様々な方法で可能であり、筋モジュールを骨格とした様々な設計を可能としている。
}%

\begin{figure}[htb]
  \centering
  \includegraphics[width=1.0\columnwidth]{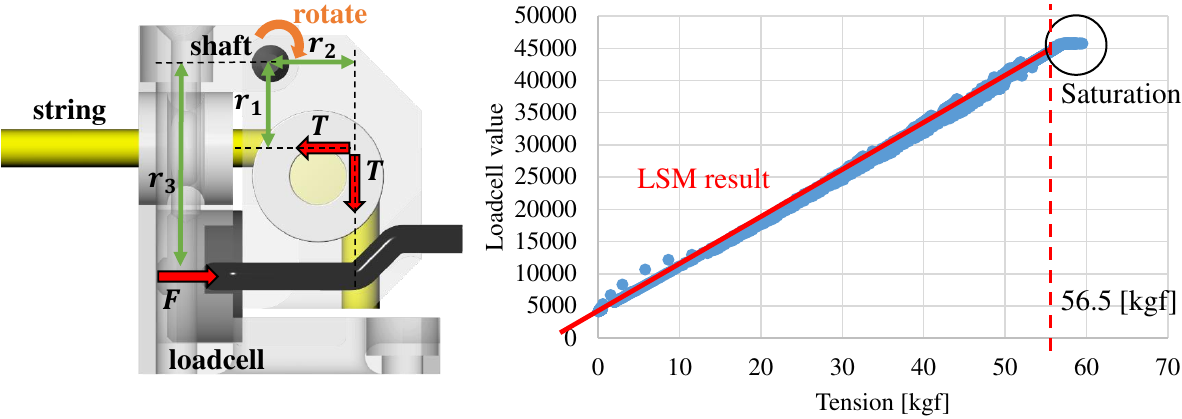}
  \vspace{-3.0ex}
  \caption{The principle of ultra tiny tension measurement unit. Left: the principle of tension measurement. Right: the result of calibration.}
  \label{figure:tension-measurement-principle}
  \vspace{-2.0ex}
\end{figure}

\switchlanguage%
{%
  Third, we discuss the ability of the ultra tiny tension measurement unit.
  The principle of tension measurement is shown to the left of \figref{figure:tension-measurement-principle}, and we will discuss the balance of moment around the shaft.
  In this study, we aim to measure muscle tension until 50 [kgf] , and set $r_1$ as 5.0 [mm], $r_2$ as 5.0 [mm], and $r_3$ as 11.3 [mm].
  By these settings, this tension measurement unit can measure tension until 56.5 [kgf] because the tension limit of the load cell $F$ is 50 [kgf] as shown in the equation below.
  \begin{align}
    T = \frac{r_3}{r_1 + r_2}F
  \end{align}
  The result of calibration is shown as the right of \figref{figure:tension-measurement-principle}, and proves that the unit can correctly measure muscle tension until 56.5 [kgf].
}%
{%
  次に、超小型張力測定ユニットの張力測定能力について議論する。
  張力測定の原理は\figref{figure:tension-measurement-principle}の左図のようになっており、軸周りのモーメントの釣り合いを考える。
  本研究においては50[kgf]までの張力を測定できることを目指し、$r_1$を5.0[mm]、$r_2$を5.0[mm]、$r_3$を11.3[mm]としている。
  これによって、ロードセルの定格が50[kgf]のため、以下の式から、56.5[kgf]まで張力が測定できることがわかる。
  \begin{align}
    T = \frac{r_3}{r_1 + r_2}F
  \end{align}
  実際にユニットのキャリブレーションを行うと\figref{figure:tension-measurement-principle}の右図のようになり、正しく56.5[kgf]までの力が測れていることがわかる。
}%

\begin{figure}[htb]
  \centering
  \includegraphics[width=1.0\columnwidth]{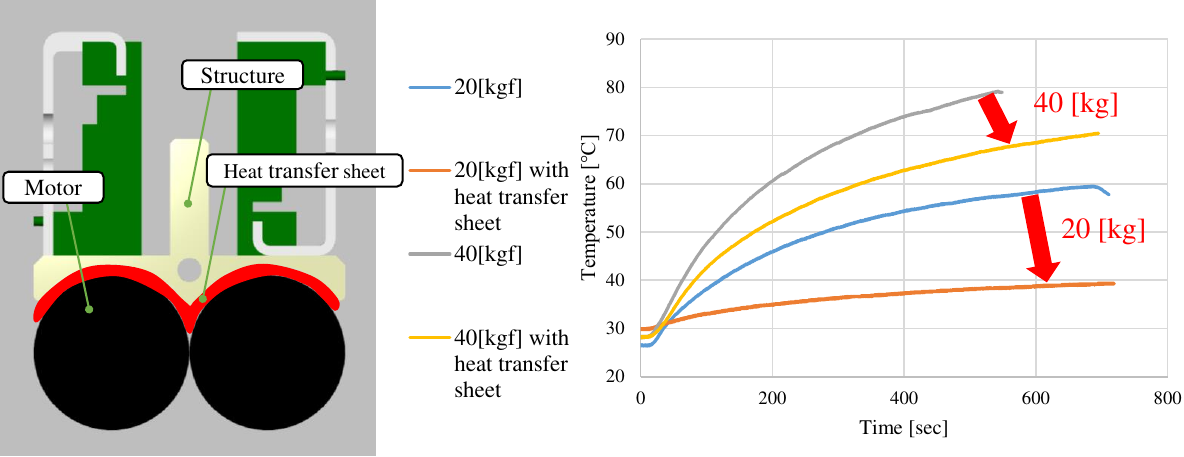}
  \caption{Comparison of motor heat transition, with and without heat transfer sheet. 20 [kgf] and 40 [kgf] weights are lifted with the newly developed miniature bone-muscle module.}
  \label{figure:double-motor-module-heat}
\end{figure}

\switchlanguage%
{%
  Fourth, we discuss the effects of suppressing the rise in temperature by dissipating motor heat to the structure.
  In this experiment, we lifted 20 [kgf] and 40 [kgf] using the muscle module, with and without insertion of the heat transfer sheet between the motor and the structure, and showed the rise in motor temperature graphically.
  We measured the temperature of the motor outer cover using the temperature sensor, and the results are shown in \figref{figure:double-motor-module-heat}.
  We can see the big suppression effect of the rise in muscle module temperature as shown in \figref{figure:double-motor-module-heat} by the dissipation of motor heat to the structure.
  This indicates that the module is able to exhibit continuously high muscle tension.
}%
{%
  次に、骨格に熱を逃がすことによる温度上昇の抑制効果について議論する。
  実験としては、ギア比157:1の筋モジュールで、伝熱性のシートを骨格とモータの間に挟んだ場合と挟まず空洞にした場合で
  20[kgf]・40[kgf]の重りを持ち上げた際の温度上昇傾向をグラフ化した。
  温度はモータ外側の温度を温度センサーで測っており、結果は\figref{figure:double-motor-module-heat}に示すようになっている。
  伝熱性のシートで骨格にモータの熱を逃がすことによって\figref{figure:double-motor-module-heat}のように、大きな熱の抑制の効果が見られる。
  よって、持続的に高張力を出すことが可能になったことが示された。
}%

\begin{figure}[htb]
  \centering
  \includegraphics[width=1.0\columnwidth]{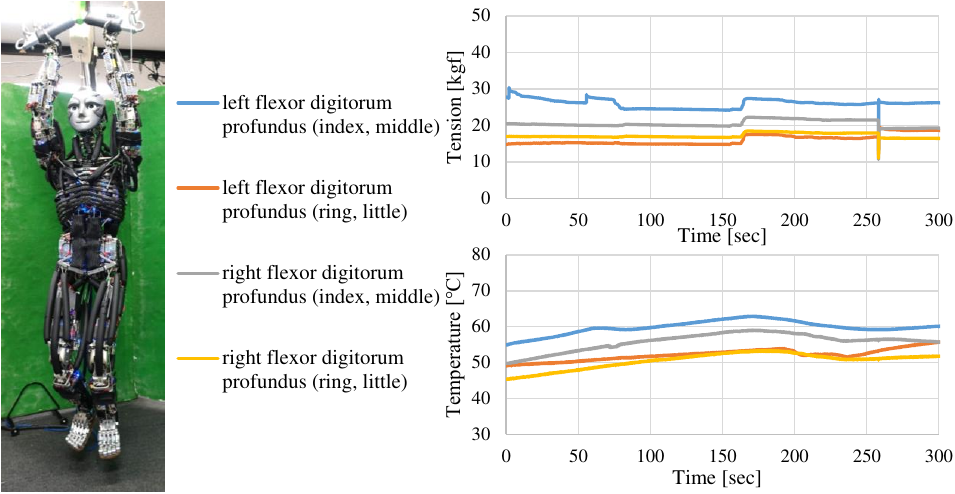}
  \caption{Result of dangling. Left: overview of dangling motion. Right: muscle tension and temperature during the experiment.}
  \label{figure:kengoro-burasagari}
\end{figure}

\switchlanguage%
{%
  Finally, we attempted to dangle Kengoro on a bar with the newly developed forearm, explained in the next section, to show that the newly developed miniature bone-muscle module functions correctly.
  We made Kengoro take the posture of dangling, fixed the muscle length, and made Kengoro dangle as shown in the right of \figref{figure:kengoro-burasagari}.
  Kengoro weighs 56 [kgf], and dangles using mainly the four left and right fingers.
  The result of muscle tension and temperature for 5 minutes is shown to the left of \figref{figure:kengoro-burasagari}.
  The tension of the muscles that actuates the fingers is 15--30 [kgf], and this temperature almost does not increase at all.
  Through this experiment, we showed the strength of the miniature bone-muscle module and its effect in inhibiting the rise of temperature.
}%
{%
  最後に、骨構造一体小型筋モジュールが正しく機能することを示すために、腱悟郎によるぶら下がり実験を試みた。
  \figref{figure:kengoro-burasagari}の左図ように姿勢を作って筋長を固定し、棒に骨構造一体小型筋モジュールによって駆動される手をかけてぶら下がらせた。
  腱悟郎の体重は56kgfであり、主に左右の手の指の筋合わせて4本を用いてぶら下がっている。
  その際の筋張力と温度は右図のようになっており、指を駆動する筋はどれも温度上昇がほとんどないことがわかる。
  このぶら下がり実験によって、骨構造一体小型筋モジュールの構造としての強さ、温度上昇による効果が示された。
}%

\section{Development of Human Mimetic Forearm with Radioulnar Joint} \label{sec:3}
\begin{figure}[htb]
  \centering
  \includegraphics[width=0.9\columnwidth]{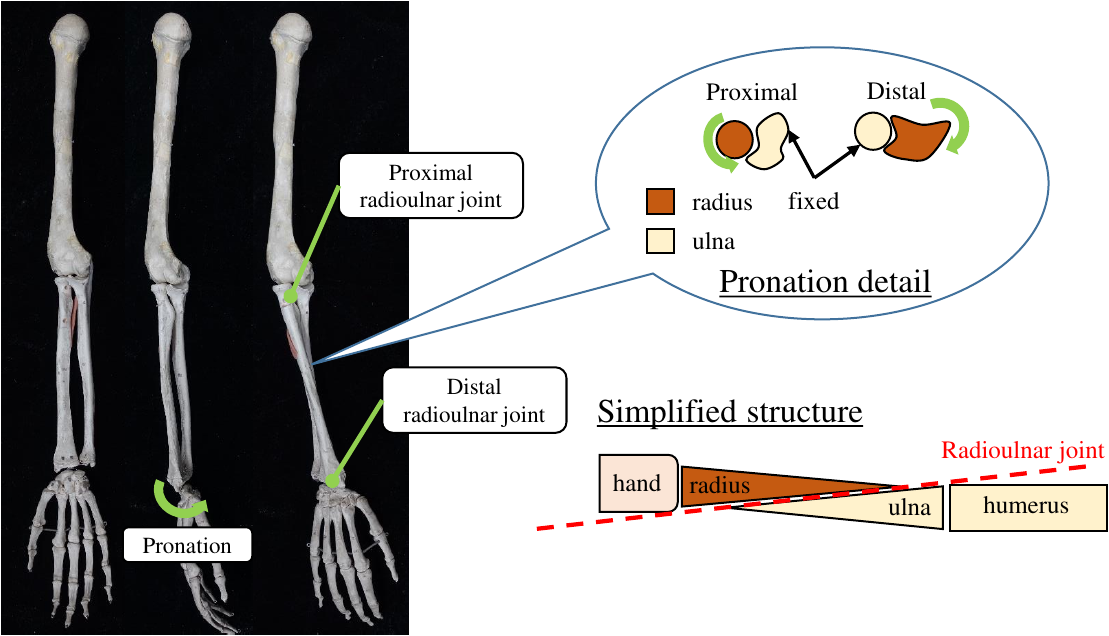}
  \caption{Structure of the human radioulnar joint.}
  \label{figure:human-radioulnar-structure}
\end{figure}

\switchlanguage%
{%
  \subsection{Human Radioulnar Structure}

  A human forearm is structured as shown in \figref{figure:human-radioulnar-structure}.
  It is composed of two long, thin bones called the radius and the ulna, and the radioulnar joint is formed by these bones and two axle joints located at the proximal and distal.
  In an ulna, the proximal is thick and the distal is thin, but in a radius, the proximal is thin and the distal is thick.
  This radioulnar structure is one of the joints that are specific to humans, and we propose its characteristics as below.
  \begin{enumerate}
    \item Even if the ulna is fixed to something completely, the radioulnar joint can move.
    \item The radioulnar joint is clinoaxis, and the joint passes the little finger through the proximal radius and the distal ulna.
    \item The radioulnar joint can disperse torsion by two long bones.
  \end{enumerate}
  As for 1), we use this characteristic when we perform motions such as writing and soldering.
  We can perform motions using 3 DOFs of the radioulnar joint and radiocarpal joint when stabilizing the arm by fixing the ulna to the table completely.
  As for 2), we use this characteristic when we perform motions such as opening a door, turning a screw, and swinging a badminton racket.
  When we open a door, we propagate torque efficiently by bending the wrist joint to the ulna and matching the axis of the radioulnar joint to the door knob joint.
  When we swing a badminton racket, we maximize the speed of the racket head by increasing the radius of rotation in bending the wrist joint to the radius and keeping the racket head away from the radioulnar joint.
  As for 3), this structure is effective for cabling and skin movements.

  We propose that these structures play a part in performing human skillful motion, and that this benefit is utilized only by imitating the body proportion, weight ratio, and muscle arrangement of the human body.
  Thus, we developed a human mimetic forearm with a radioulnar joint using newly developed miniature bone-muscle modules.
}%
{%
  \subsection{人体の橈骨尺骨構造}
  人間の前腕は\figref{figure:human-radioulnar-structure}のような構造と成っている。
  橈骨・尺骨と呼ばれる二本の骨によって前腕は構成され、これらの近位と遠位にある車軸関節によって橈尺関節を成している。
  尺骨は近位が太く遠位が細いのに対し、橈骨は近位が細く遠位が太い。
  そして、上腕は主に尺骨に接続し、手は主に橈骨に接続している。
  この橈骨尺骨構造は人間特有な関節の一つであるが、特徴としては以下が挙げられると考えている。
  \begin{enumerate}
    \item 尺骨を地面などに固定した状態でも橈骨のみによって橈尺関節が動作する。
    \item 関節軸が斜めに入っており、近位橈骨から遠位尺骨を通って小指辺りを通過している。
    \item 二本の長い骨によってねじりを分散することができる。
  \end{enumerate}
  1)としては、文字を書いたり、ハンダ付けをする際にこの特徴が用いられる。
  尺骨をべったりとテーブルに固定して安定させ、橈尺関節と橈骨手根関節の3自由度で動作を行うことができる。
  2)としては、ドア開けやねじ回し、バドミントンをする際にこの特徴が用いられる。
  ドア開けは、手首を尺骨側に曲げてドアノブの軸と橈尺関節の軸とを一致させて効率的なトルク伝達を行う。
  バドミントンは手首を橈骨側に曲げることでラケットヘッドを橈尺関節の軸から離し、ヘッドのスイングスピードを最大化する。
  3)としては、ケーブリングや皮膚などに関して有用であると考えられる。
  これらのように、この構造は人体の巧みな動作の実現に一役買っていると考え、そして人体のプロポーションや重量、筋配置などを模倣することによって初めてその利点が生かされると考える。
  そこで我々は、本研究で開発した骨構造一体小型筋モジュールを用いて、人体模倣型橈骨尺骨構造を有する前腕部の設計を行った。
}%

\begin{figure}[htb]
  \centering
  \includegraphics[width=0.8\columnwidth]{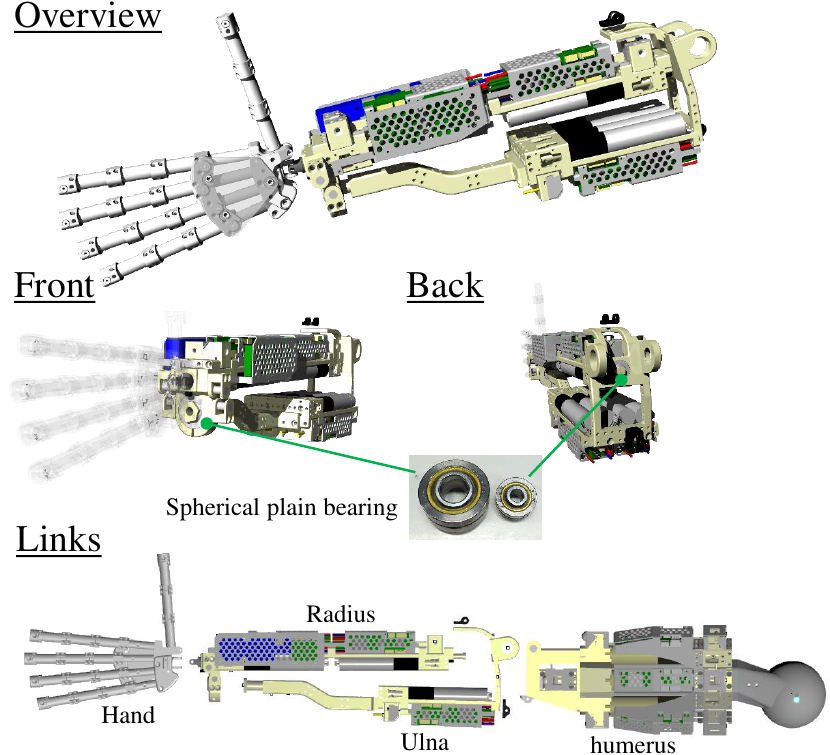}
  \caption{Overview of newly developed Kengoro forearm.}
  \label{figure:kengoro-radioulnar-structure}
\end{figure}

\switchlanguage%
{%
  \subsection{Realization of Human Mimetic Radioulnar Structure}
  The developed forearm with a radioulnar joint is shown in \figref{figure:kengoro-radioulnar-structure}.
  It is very compact, enabled by making most of the benefit that the miniature bone-muscle module is able to connect lengthwise and crosswise to form the structure.
  Two modules each are equipped in the radius and ulna, and the radius is almost completely composed of only modules.
  There are 4 modules in total, and thus 8 muscles, in the forearm.
  The radius is thick at the distal like that of a human, and connects to the hand \cite{iros2017:makino:hand} through a universal joint.
  Likewise, the ulna is thick at the proximal, and connects to the humerus.
  To rotate the radioulnar joint, spherical plain bearings are equipped in the proximal of the radius and the distal of the ulna as axle joints.
}%
{%
  \subsection{人体模倣型橈骨尺骨構造の実現方法}
  本論文で開発した橈骨尺骨構造を有する前腕は\figref{figure:kengoro-radioulnar-structure}のようになっており、骨構造一体小型筋モジュールの、骨格として縦にも横にも接続できるとうい利点を活かしたコンパクトな作りとなっている。
  橈骨、尺骨にそれぞれ二つずつの骨構造一体小型筋モジュールが搭載されており、橈骨については完全に筋モジュールのみで骨格が構成されている。
  合わせて筋モジュールが4つ、つまり8つの筋が前腕部に搭載されている。
  人体と同じように、橈骨は遠位で太く、その橈骨に手\cite{makino's hand}がユニバーサルジョイントを介して接続している。
  同様に、尺骨は近位で太く、その尺骨が上腕に接続している。
  近位橈尺関節、遠位橈尺関節には球面滑り軸受けが搭載されており、これによって橈尺関節の回転を行う。
}%

\begin{figure}[b]
  \centering
  \includegraphics[width=1.0\columnwidth]{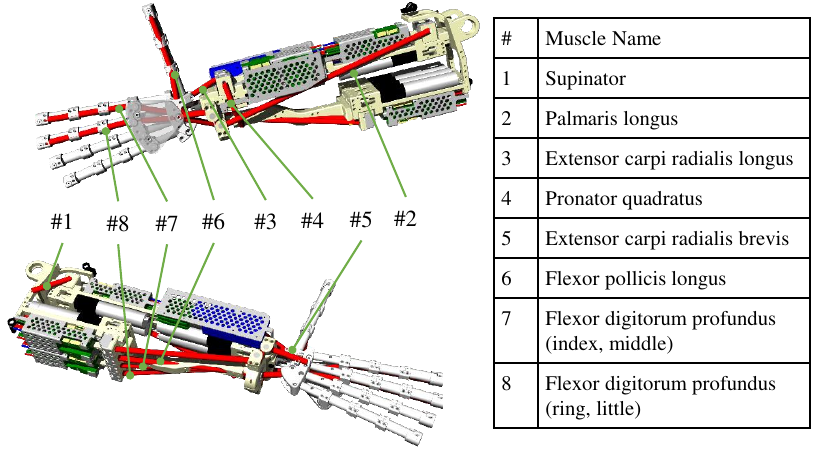}
  \vspace{-3.0ex}
  \caption{Muscle arrangement of the newly developed forearm.}
  \label{figure:kengoro-radioulnar-muscle}
\end{figure}

\begin{figure}[htb]
  \centering
  \includegraphics[width=0.7\columnwidth]{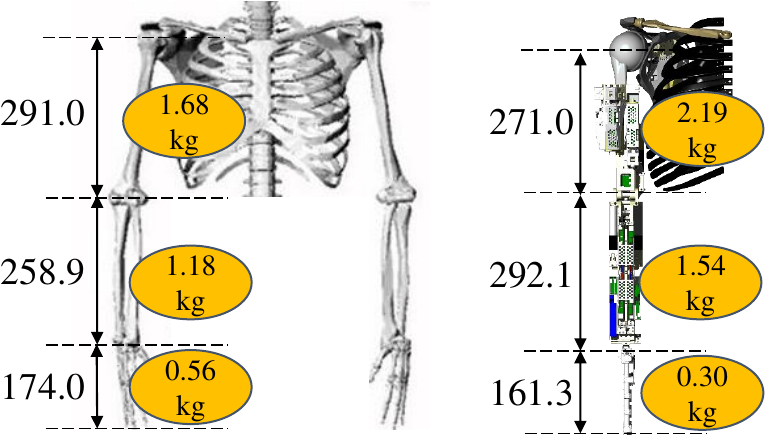}
  \caption{Comparison of upper limb link length and weight between a human and Kengoro with a newly developed forearm.}
  \label{figure:kengoro-human-property}
\end{figure}

\switchlanguage%
{%
  \subsection{Performance of Developed Forearm}
  The muscle arrangement is shown in \figref{figure:kengoro-radioulnar-muscle}.
  We imitated 8 muscles in the human forearm, and there are 6 DOFs that are moved by the 8 muscles, including 1 DOF of the radioulnar joint, 2 DOFs of the radiocarpal joint and 3 DOFs of the fingers (thumb, index and middle, ring and little).
  In these muscles, the gear ratios of $\#1$, $\#4$ and $\#6$ are 84:1, and those of the others are 157:1.
  The number of muscles can be an important index in expressing how much freedom the forearm has, and this forearm actualizes many more muscles compactly compared to other robots such as Anthrob \cite{humanoids2013:michael:anthrob} (2 muscles), Kenshiro \cite{humanoids2012:nakanishi::kenshirodesign} (0 muscles), and Kenzoh \cite{robio2011:nakanishi:kenzoh} (5 muscles).
  Also, we succeeded in imitating the human body without deviating from the human body proportion and weight ratio as shown in \figref{figure:kengoro-human-property}.
  We show the workspace and maximum torque of 4 DOFs of the elbow joint, radioulnar joint, and radiocarpal joint developed in this study in \tabref{table:comparison-joint-performance}.
  This also indicates that the forearm is correctly based on the human body.
  Thus, we succeeded in developing a forearm with a radioulnar joint, which has many degrees of freedom and is based on the body proportion, weight ratio, muscle arrangement, and joint performance of the human body.
}%
{%
  \subsection{筋配置とプロポーションと性能}
  筋配置は\figref{figure:kengoro-radioulnar-muscle}にあるような形となっている。
  手の8つの筋肉を模し、自由度としては、橈尺関節の1自由度、手首関節の2自由度、指は親指、人差し指、中指・薬指・小指の3自由度を8つの筋で動作させている。
  このうち$\#1$ $\#4$ $\#6$はギア比が84:1であり、他は157:1である。
  前腕に含まれる筋数は指や手首の多自由度を表現するための重要な指標になりうるが、\cite{humanoids2013:michael:anthrob}では2、
  \cite{humanoids2012:nakanishi::kenshirodesign}では0、\cite{robio2011:nakanishi:kenzoh}は5などに比べるとより多くの筋をコンパクトに実現できていることがわかる。
  また、プロポーションや重量に関しては\figref{figure:kengoro-human-property}のように逸脱しない形で人体を模倣することに成功している。
  そして、本研究で開発した肘関節、橈尺関節、橈骨手根関節の4自由度に関して可動域と最大トルクを\tabref{table:comparison-joint-performance}に示すが、
  これもまた正しく人体を模倣できていることがわかる。
  よって、骨構造一体小型筋モジュールというアプローチによって今まで実現していなかった、多自由度であり、人体のプロポーションや重量、関節性能を模倣した橈骨尺骨構造を
  有する前腕の開発に成功した。
}%

\begin{table}[htb]
  \centering
  \caption{Comparison between joint performance of a human and that of Kengoro.}
  \scalebox{0.90}{
    \begin{tabular}{l r |r r|r r }
      \hline
      & & \multicolumn{2}{c|}{Human${^*}{^1}$} & \multicolumn{2}{|c}{Kengoro} \\ \hline
      Joint & & Torque & Workspace & Torque${^*}{^2}$ & Workspace\\
            & & [Nm]  & [deg] & [Nm] & [deg] \\
      \hline \hline
      Elbow
      & pitch & -72.5 -- 42.1 & -145 -- 0 & -49.9 -- 46.5 & -145 -- 0\\
      Radioulnar
      & yaw & -7.3 -- 9.1  & -90 -- 85 &  -8.5 -- 3.3 & -85 -- 85\\
      Wrist
      & roll & -12.2 -- 7.1 & -85 -- 85 & -15.1 -- 14.6 & -75 -- 85\\
      & pitch & -11 -- 9.5 & -15 -- 45 & -15.9 -- 13.3 & -15 -- 45\\
      \hline
      \multicolumn{6}{l}{${^*}{^1}$ \cite{none:kapandji:upper, none:neiman:musclekinesiology}} \\
      \multicolumn{6}{l}{${^*}{^2}$ simulated value} \\
    \end{tabular}
  }
  \label{table:comparison-joint-performance}
\end{table}

\section{Achievement of Human Skillful Motion using Radioulnar Structure} \label{sec:4}
\begin{figure}[htb]
  \centering
  \includegraphics[width=1.0\columnwidth]{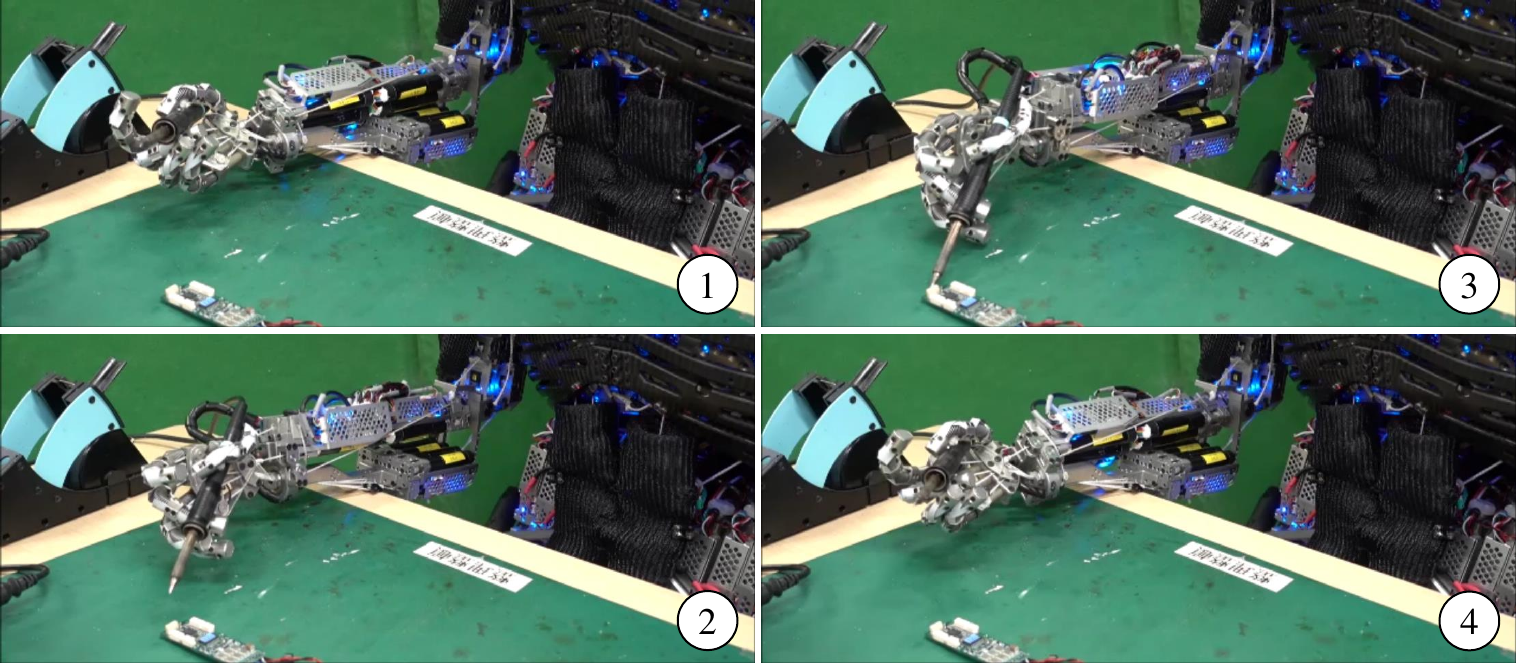}
  \vspace{-3.0ex}
  \caption{Kengoro soldering. Kengoro with a soldering iron can move the radioulnar joint with the ulna attached to the table.}
  \label{figure:kengoro-soldering}
\end{figure}

\switchlanguage%
{%
  Due to the success in the development of a radioulnar structure based on the human body proportion, we propose that Kengoro is able to move in various ways using the benefits of this radioulnar structure.
  Thus, we performed some human-specific motions using Kengoro \cite{humanoids2016:asano:kengoro} equipped with the forearm having the radioulnar joint.
  In this section, we will evaluate the degree of imitation of the forearm and verify the benefits of the radioulnar structure through experiments conducted on motion that uses the benefits described in the previous chapter, such as soldering, opening a book, turning a screw, and swinging a badminton racket.

  \subsection{Soldering}
  The motion of soldering (\figref{figure:kengoro-soldering}) is an example that effectively uses the characteristic that the radioulnar joint can move even with the ulna attached to something.
  We can see that Kengoro is able to move the radioulnar joint stably with the ulna attached to the table.
  This characteristic is thought to also be seen when writing and using a keyboard.
  Typically, large and strong structures are needed in order to make robots with high rigidity for stable hand movement.
  However, if the robot has a low rigidity, stable and fine movements can be done by having a radioulnar joint and moving the radioulnar and radiocarpal joints with the ulna bone attached to something.
  We propose that this can support the drawback of being unable to do fine movements by the tendon-driven musculoskeletal humanoid, which has safe structures but low rigidity.
}%
{%
  人体プロポーションに合った橈骨尺骨構造を有する前腕の開発に成功したことにより、この利点を用いた様々な動作ができると考える。
  そこで、橈骨尺骨構造を有する前腕を搭載した腱悟郎\cite{humanoids2016:asano:kengoro}によって橈骨尺骨構造だからこそできる人間特有の動作を行う。
  本研究では、前節で説明したうちの二つの特徴を活かしたハンダ付け動作、本開け動作、ネジ回し動作、バドミントン動作を行うことで、前腕の人体模倣度の検証、橈骨尺骨構造の利点の検証を行う。

  \subsection{ハンダ付け動作}
  ハンダ付け動作(\figref{figure:kengoro-soldering})は、橈尺関節を動作させる際に尺骨は動かず橈骨のみが動くという特徴を上手く使った例である。
  尺骨を机に固定しながら安定して橈尺関節を動作できる。
  この動作は文字を書くときやキーボードを打つときにも見られると考える。
  一般的に、手先がぶれないようにロボットを高剛性にするためには強く大きな構造が必要となる。
  しかし、橈骨尺骨構造を有することによって、低剛性でも尺骨側をテーブルなどにべったりとつけたまま橈尺関節と橈骨手根関節によって細かい動作を行うことができる。
  これは低剛性で安全な筋骨格腱駆動ヒューマノイドによる細かい動作に対するビハインドを補うことができると考える。
}%

\begin{figure}[htb]
  \centering
  \includegraphics[width=1.0\columnwidth]{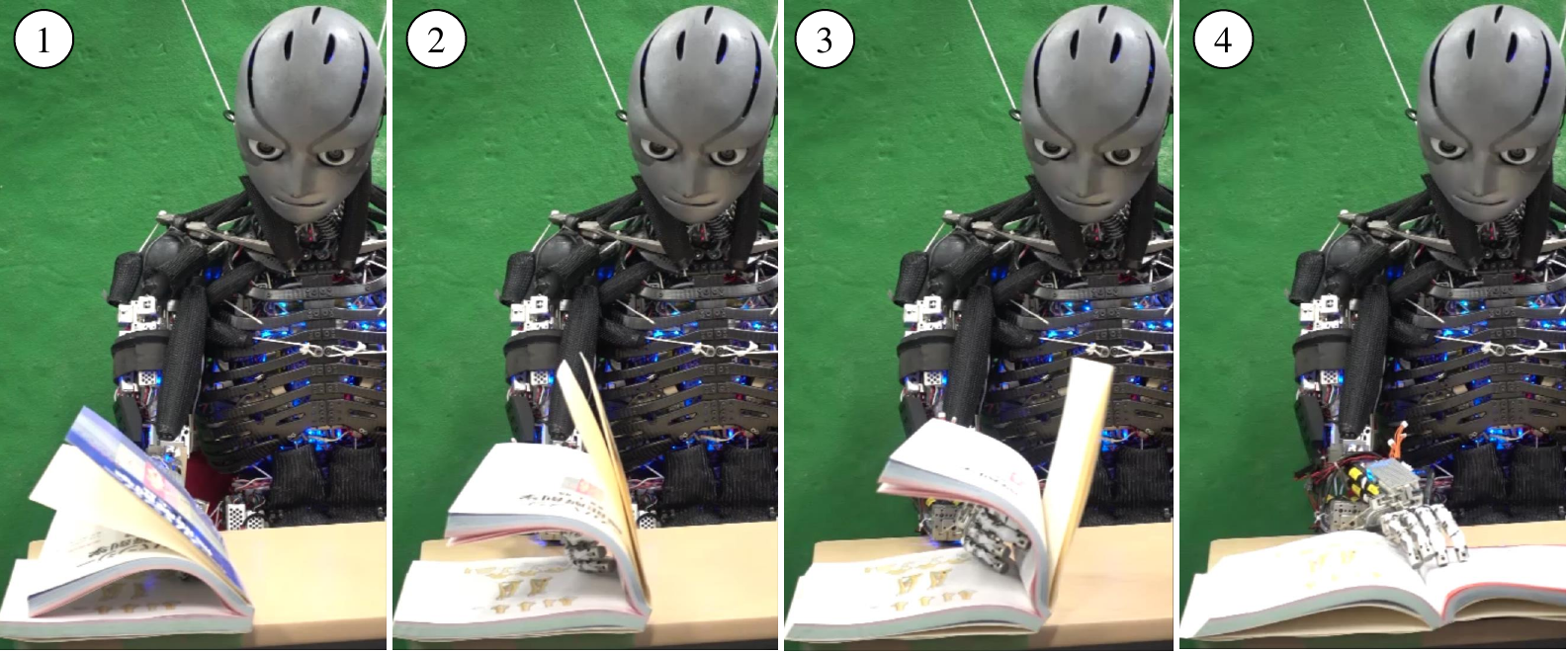}
  \vspace{-3.0ex}
  \caption{Kengoro opening a book.}
  \label{figure:kengoro-book}
\end{figure}

\begin{figure}[htb]
  \centering
  \includegraphics[width=1.0\columnwidth]{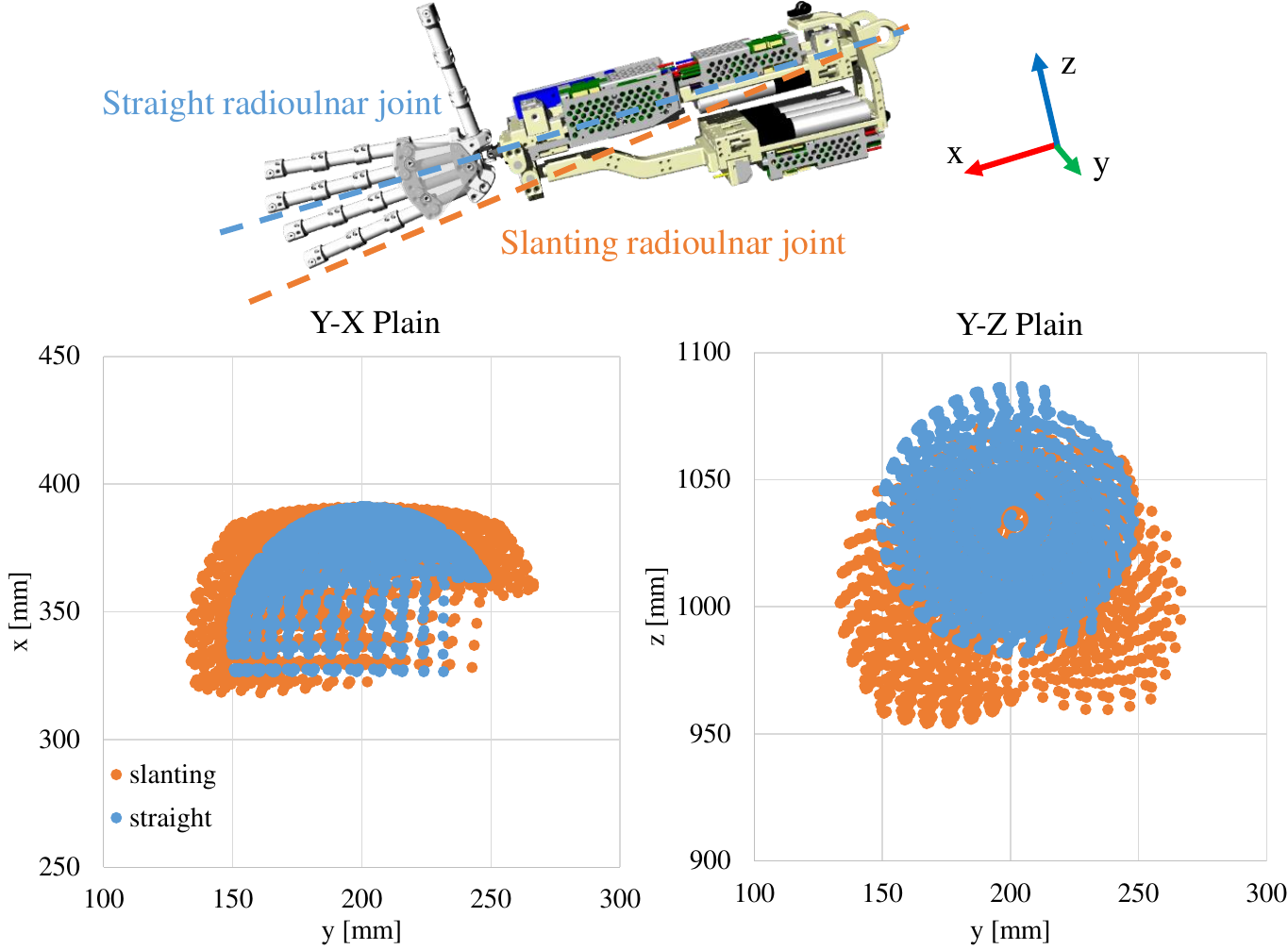}
  \vspace{-3.0ex}
  \caption{The reachable points of the center of the palm compared between the slanting radioulnar joint and the ordinary straight radioulnar joint. Left: x-y plain. Right: y-z plain.}
  \label{figure:kengoro-radioulnar-reachability}
\end{figure}

\switchlanguage%
{%
  \subsection{Opening a Book}
  The motion of opening a book (\figref{figure:kengoro-book}) is an example that effectively uses the characteristic that the radioulnar joint axis is slanting and passes through at about the little finger.
  We can see that Kengoro is able to open a book by merely rotating the radioulnar joint, which becomes a motion like that of turning the palm.
  Also, we can say that this extends the capacity of movement.
  \figref{figure:kengoro-radioulnar-reachability} is the comparison between an ordinary straight radioulnar joint and the slanting radioulnar joint of the reachable points of the center of the palm, that can be reached by only using the radioulnar and radiocarpal joints.
  The slanting radioulnar joint can extend hand movement, and the hand can move widely and stably by combining this and the previous benefit that the radioulnar joint can move even with the ulna attached to something.
}%
{%
  \subsection{本開け動作}
  本開け動作(\figref{figure:kengoro-book})は橈尺関節軸が斜めになっており、小指辺りを通過していることを上手く使った例である。
  単純に橈尺関節を回転させるだけで手のひらを返すような動作になり、本を開けることができていることがわかる。
  また、これは動作範囲の拡張と言うこともできる。
  \figref{figure:kengoro-radioulnar-reachability}は、橈尺関節が斜めの場合と真っ直ぐの場合において、手のひらの中心点が橈尺関節と橈骨手根関節のみを使って動作できる範囲を比べたものである。
  斜めの軸を持つことによって手の動作範囲が拡張されており、これと前に説明したような尺骨側を何かしらに完全にくっつけた状態でも橈尺関節が動作できるという利点を組み合わせることで、安定した状態で広く手を動作することができるのである。
}%

\begin{figure}[htb]
  \centering
  \includegraphics[width=1.0\columnwidth]{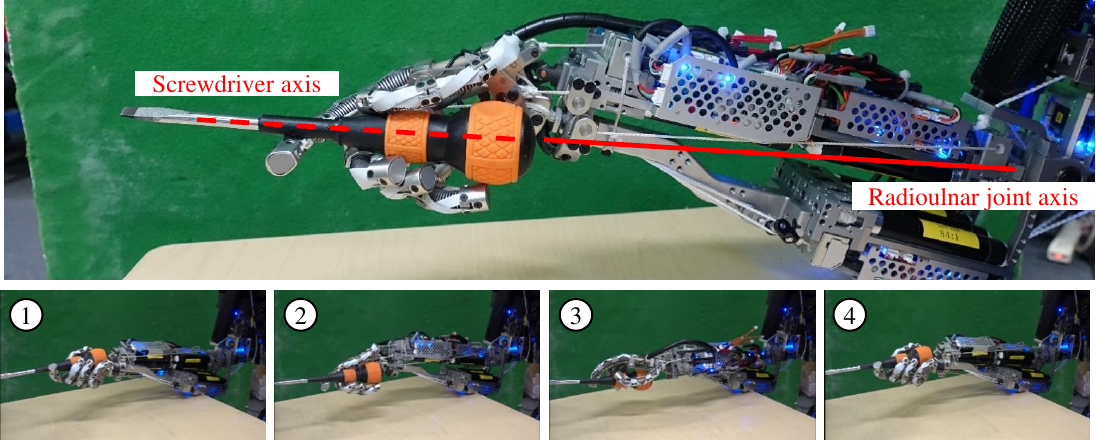}
  \vspace{-3.0ex}
  \caption{Kengoro turning a screw with a screwdriver. Upper picture shows that the radioulnar joint axis matches the screwdriver axis.}
  \label{figure:kengoro-neji}
\end{figure}

\switchlanguage%
{%
  \subsection{Turning a Screw}
  When turning a screw with a screwdriver, Kengoro can transfer torque efficiently by matching the radioulnar joint axis to the axis of the screwdriver.
  We can see that the tip of the screwdriver is hardly blurred.
  The motion of opening a door uses the same principle.
}%
{%
  \subsection{ネジ回し動作}
  ネジ回し動作(\figref{figure:kengoro-neji})は斜めに入った橈尺関節軸とドライバーの軸を一致させることによってトルクを伝達している。
  橈尺関節を回転させてもドライバーの先端がほとんどブレていないことがわかる。
  ドアを開ける際も同様の原理である。
}%

\begin{figure}[htb]
  \centering
  \includegraphics[width=1.0\columnwidth]{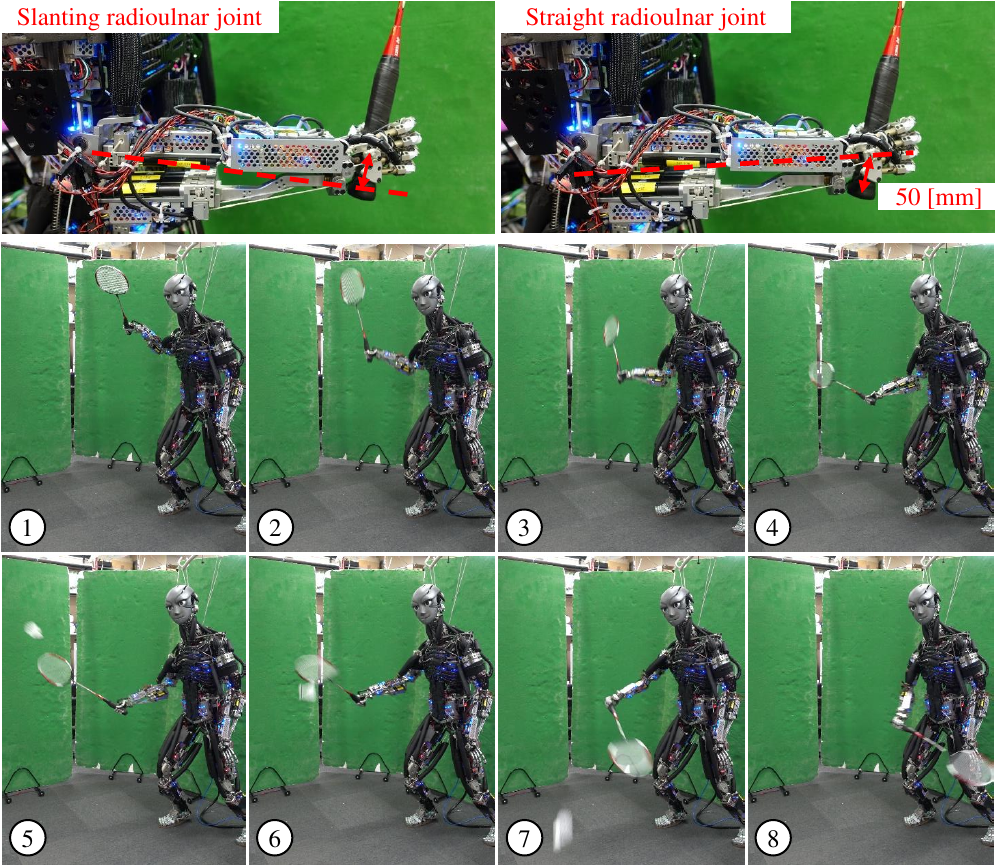}
  \vspace{-3.0ex}
  \caption{Badminton swing motion. Upper pictures show comparison between the slanting radioulnar structure with large radius of rotation of racket and the ordinary straight radioulnar structure with small radius of rotation of racket.}
  \label{figure:kengoro-badminton}
\end{figure}

\begin{figure}[htb]
  \centering
  \includegraphics[width=1.0\columnwidth]{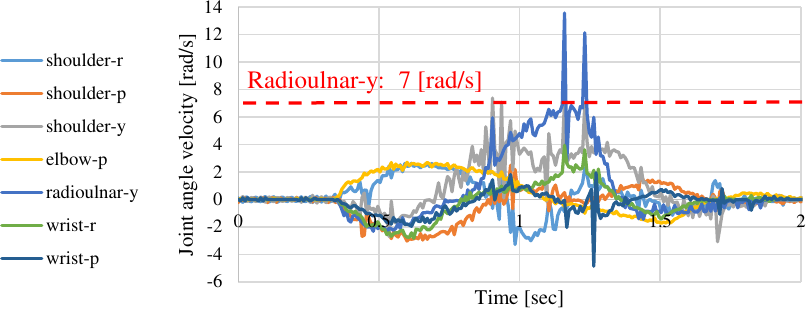}
  \vspace{-3.0ex}
  \caption{Joint angle velocity of badminton swing motion.}
  \label{figure:kengoro-badminton-log}
\end{figure}

\switchlanguage%
{%
  \subsection{Badminton Swing}
  When swinging a badminton racket (\figref{figure:kengoro-badminton}), Kengoro can increase the radius of rotation and speed in the racket head by keeping the hand away from the radioulnar joint.
  Due to the slanting radioulnar joint, Kengoro can have a larger radius of rotation than with the ordinary straight radioulnar joint.
  This motion contrasts with the motion of turning a screw, and is a skillful human movement that uses the effects of the slanting radioulnar joint for speed of the swing instead of the torque.
  In this study, we used the optimization method of \cite{humanoids2016:terasawa:optimization} to create the badminton swing motion, and made Kengoro move in this way.
  The joint angle velocity of Kengoro during this motion is shown in \figref{figure:kengoro-badminton-log}, and the speed of the radioulnar joint was the fastest.
  Specifically, the slanting radioulnar joint increases the radius of rotation of racket by about 50 [mm] compared with the ordinary straight radioulnar joint, and the increase of the racket speed by the slanting joint is 0.35 [m/s] in contrast to the total racket speed of 8 [m/s], thus the effect is about 4.3 [\%].
  This is not a big effect, but shows that the radioulnar joint is important in competitive sports that require speed, and is very important to be used properly and skillfully.
}%
{%
  \subsection{スイング動作}
  バドミントンのスイング動作(\figref{figure:kengoro-badminton})は斜めに入った橈尺関節軸からなるべくラケットのヘッドを遠ざけることによってヘッドのスピードを稼いでいる。
  この動作はネジ回し動作とは対照的な動作であり、斜軸の効果をトルクではなく速度に振り分けるという人間らしい動作である。
  本研究では\cite{humanoids2016:terasawa:optimization}の手法を用いてバドミントン動作を生成し、腱悟郎を動作させている。
  この際の腱悟郎の関節スピードは\figref{figure:kengoro-badminton-log}のようになっており、橈尺関節のスピードが最も大きい。
  具体的には、通常の真っ直ぐな橈尺関節に比べ斜めの橈尺関節ではラケットヘッドの回転半径を約50[mm]増やすことができているため、この動作の際のヘッドのスピードである8[m/s]に対して橈尺関節の斜め軸によるスピード上昇の効果は0.35[m/s]であり、約4.3[\%]程度である。
  これは大きな効果ではないが、橈尺関節がバドミントンのようなスピードを必要とする競技に非常に重要であると同時に、橈尺関節の斜軸による巧みな使い分けが非常に重要であることがわかる。
}%

\section{CONCLUSION} \label{sec:5}
\switchlanguage%
{%
  In this study, we explained the development of the human mimetic forearm with a radioulnar joint made by miniature bone-muscle modules.
  First, we explained the need for a forearm with a radioulnar joint that is based on the body proportion, weight ratio, and muscle arrangement of the human body in order to achieve human skillful motion.
  Then, we explained the need for the miniaturization of muscle modules to save space in order to actualize the human mimetic radioulnar joint of the tendon-driven musculoskeletal humanoid.
  To approach this, we proposed the method of using space efficiently by installing two muscle actuators in one muscle module, integrating muscle and bone structure, and using a more miniature motor and solving its drawbacks by dissipating motor heat to the structure.
  We succeeded in developing a forearm that is based on the body proportion, weight ratio, muscle arrangement, and joint performance of the human body using newly developed miniature bone-muscle modules.
  Finally, we conducted experiments on some motions using characteristics of the radioulnar joint, such as the ability to move with the ulna attached to something, and that the joint is slanting.
  Through these experiments, we proposed the correctness of the approach in the human mimetic radioulnar joint with miniature bone-muscle modules, and observed the benefits of the radioulnar joint.

  For future works, we propose the actualization of a small tendon-driven musculoskeletal humanoid made of the newly developed miniature bone-muscle modules.
  These miniature bone-muscle modules can be used for the forearm, as well as various other parts of the tendon-driven robot.
  At the same time, we aim to understand the biological meaning of the radioulnar joint, and find motions that use this joint that are more skillful.
}%
{%
  本論文では、人体の橈骨尺骨構造を模した前腕のための骨構造一体小型筋モジュールの開発について述べた。
  まず、上肢の巧みな動作実現のためには人体のプロポーションや重量・筋配置などを模倣した橈骨尺骨構造を有する前腕が必要なことを述べた。
  そして、筋骨格腱駆動ヒューマノイドにおいて人体模倣型の橈骨尺骨構造を実現するためには筋モジュールの小型化・省スペース化が必要なことを述べ、そのためのアプローチとして、
  1モジュール2筋として空間を有利に使い、骨構造と筋を一体化すること、また小型モータを使い、その出力的ビハインドを骨格に熱を逃がすことで解決する方法を述べた。
  この骨構造一体小型筋モジュールを用いることで、今まで達成し得なかった、人体を模したプロポーションや重量・筋配置の橈骨尺骨構造を有する前腕を実現することを可能とした。
  最後に、橈骨尺骨構造の利点であると考えられる、内旋外旋時に尺骨が動かないこと、橈尺関節軸が斜軸となっていることを利用した複数の動作を行い、
  骨構造一体小型筋モジュールによる人体プロポーションに合った橈骨尺骨構造というアプローチの正しさと、橈骨尺骨構造の有用性を見出した。

  今後としては、本論文における骨構造一体小型筋モジュールを用いることでより小型の筋骨格腱駆動ヒューマノイドの開発を実現できると考えている。
  この骨構造一体小型筋モジュールというアプローチは前腕部のみに適用可能なものではなく、広く様々な部位に使用することができると考える。
  また同時に、この橈骨尺骨構造の生物学的意味の理解や、橈骨尺骨構造を用いたより巧みな動作例の発見も目指したい。
}%

{
  \bibliographystyle{IEEEtran}
  \bibliography{main}

\begin{thebibliography}{10}
\providecommand{\url}[1]{#1}
\csname url@rmstyle\endcsname
\providecommand{\newblock}{\relax}
\providecommand{\bibinfo}[2]{#2}
\providecommand\BIBentrySTDinterwordspacing{\spaceskip=0pt\relax}
\providecommand\BIBentryALTinterwordstretchfactor{4}
\providecommand\BIBentryALTinterwordspacing{\spaceskip=\fontdimen2\font plus
\BIBentryALTinterwordstretchfactor\fontdimen3\font minus
  \fontdimen4\font\relax}
\providecommand\BIBforeignlanguage[2]{{%
\expandafter\ifx\csname l@#1\endcsname\relax
\typeout{** WARNING: IEEEtran.bst: No hyphenation pattern has been}%
\typeout{** loaded for the language `#1'. Using the pattern for}%
\typeout{** the default language instead.}%
\else
\language=\csname l@#1\endcsname
\fi
#2}}

\bibitem{icra1998:hirai:asimo}
K.~Hirai, M.~Hirose, Y.~Haikawa, and T.~Takenaka, ``{The Development of Honda
  Humanoid Robot},'' in \emph{Proceedings of The 1998 IEEE International
  Conference on Robotics and Automation}, 1998, pp. 1321--1326.

\bibitem{humanoids2010:hugo:eccerobot}
H.~G. Marques, M.~J{\"a}ntsh, S.~Wittmeier, O.~Holland, C.~Alessandro,
  A.~Diamond, M.~Lungarella, and R.~Knight, ``{ECCE1: the first of a series of
  anthropomimetic musculoskeletal upper torsos},'' in \emph{Proceedings of the
  2010 IEEE-RAS International Conference on Humanoid Robots}, 2010, pp.
  391--396.

\bibitem{humanoids2016:asano:kengoro}
Y.~Asano, T.~Kozuki, S.~Ookubo, M.~Kawamura, S.~Nakashima, T.~Katayama,
  Y.~Iori, H.~Toshinori, K.~Kawaharazuka, S.~Makino, Y.~Kakiuchi, K.~Okada, and
  M.~Inaba, ``{Human Mimetic Musculoskeletal Humanoid Kengoro toward Real World
  Physically Interactive Actions},'' in \emph{Proceedings of the 2016 IEEE-RAS
  International Conference on Humanoid Robots}, 2016, pp. 876--883.

\bibitem{iros2012:asano:knee}
Y.~Asano, H.~Mizoguchi, T.~Kozuki, Y.~Motegi, M.~Osada, J.~Urata, Y.~Nakanishi,
  K.~Okada, and M.~Inaba, ``{Lower Thigh Design of Detailed Musculoskeletal
  Humanoid Kenshiro},'' in \emph{Proceedings of the 2012 IEEE/RSJ International
  Conference on Intelligent Robots and Systems}, 2012, pp. 4367--4372.

\bibitem{iros2007:sodeyama:kojiro-shoulder}
Y.~Sodeyama, T.~Yoshikai, T.~Nishino, I.~Mizuuchi, and M.~Inaba, ``{The Designs
  and Motions of a Shoulder Structure with a Wide Range of Movement Using
  Bladebone-Collarbone Structures},'' in \emph{Proceedings of the 2007 IEEE/RSJ
  International Conference on Intelligent Robots and Systems}, 2007, pp.
  3629--3634.

\bibitem{iros2012:ikemoto:shoulder}
S.~Ikemoto, F.~Kannou, and K.~Hosoda, ``{Humanlike Shoulder Complex for
  Musculoskeletal Robot Arms},'' in \emph{Proceedings of the 2012 IEEE/RSJ
  International Conference on Intelligent Robots and Systems}, 2012, pp.
  4892--4897.

\bibitem{robio2011:nakanishi:kenzoh}
Y.~Nakanishi, T.~Izawa, M.~Osada, N.~Ito, S.~Ohta, J.~Urata, and M.~Inaba,
  ``{Development of Musculoskeletal Humanoid Kenzoh with Mechanical Compliance
  Changeable Tendons by Nonlinear Spring Unit},'' in \emph{Proceedings of the
  2011 IEEE International Conference on Robotics and Biomimetics}, 2011, pp.
  2384--2389.

\bibitem{iros2015:asano:module}
Y.~Asano, T.~Kozuki, S.~Ookubo, K.~Kawasaki, T.~Shirai, K.~Kimura, K.~Okada,
  and M.~Inaba, ``{A Sensor-driver Integrated Muscle Module with High-tension
  Measurability and Flexibility for Tendon-driven Robots},'' in
  \emph{Proceedings of the 2015 IEEE/RSJ International Conference on
  Intelligent Robots and Systems}, 2015, pp. 5960--5965.

\bibitem{humanoids2013:michael:anthrob}
M.~J{\"a}ntsch, S.~Wittmeier, K.~Dalamagkidis, A.~Panos, F.~Volkart, and
  A.~Knoll, ``{Anthrob - A Printed Anthropomimetic Robot},'' in
  \emph{Proceedings of the 2013 IEEE-RAS International Conference on Humanoid
  Robots}, 2013, pp. 342--347.

\bibitem{humanoids2007:mizuuchi:kojiro}
I.~Mizuuchi, Y.~Nakanishi, Y.~Sodeyama, Y.~Namiki, T.~Nishino, N.~Muramatsu,
  J.~Urata, K.~Hongo, T.~Yoshikai, and M.~Inaba, ``{An Advanced Musculoskeletal
  Humanoid Kojiro},'' in \emph{Proceedings of the 2007 IEEE-RAS International
  Conference on Humanoid Robots}, 2007, pp. 294--299.

\bibitem{humanoids2012:nakanishi::kenshirodesign}
Y.~Nakanishi, Y.~Asano, T.~Kozuki, H.~Mizoguchi, Y.~Motegi, M.~Osada,
  T.~Shirai, J.~Urata, K.~Okada, and M.~Inaba, ``{Design Concept of Detail
  Musculoskeletal Humanoid Kenshiro -Toward a real human body musculoskeletal
  simulator-},'' in \emph{Proceedings of the 2012 IEEE-RAS International
  Conference on Humanoid Robots}, 2012, pp. 1--6.

\bibitem{humanoids2016:johannes:toro}
J.~Englsberger, A.~Werner, C.~Ott, B.~Henze, M.~A. Roa, G.~Garofalo, R.~Burger,
  A.~Beyer, O.~Eiberger, K.~Schmid, and A.~Albu-Schaffer, ``{Overview of the
  torque-controlled humanoid robot TORO},'' in \emph{Proceedings of the 2014
  IEEE-RAS International Conference on Humanoid Robots}, 2014, pp. 916--923.

\bibitem{iros2017:makino:hand}
S.~Makino, K.~Kawaharazuka, M.~Kawamura, Y.~Asano, K.~Okada, and M.~Inaba,
  ``High-power, flexible, robust hand: Development of musculoskeletal hand
  using machined springs and realization of self-weight supporting motion with
  humanoid, in press,'' in \emph{Proceedings of the 2017 IEEE/RSJ International
  Conference on Intelligent Robots and Systems}, 2017.

\bibitem{none:kapandji:upper}
I.A.KAPANDJI, \emph{PHYSIOLOGIE ARTICULAIRE}, 6th~ed.\hskip 1em plus 0.5em
  minus 0.4em\relax Ishiyaku Pub,Inc, 2010, vol.~1.

\bibitem{none:neiman:musclekinesiology}
D.~A. Neumann, \emph{Kinesiology of the Musculoskeletal System: Foundations for
  Rehabilitation}.\hskip 1em plus 0.5em minus 0.4em\relax Mosby, 2013.

\bibitem{humanoids2016:terasawa:optimization}
R.~Terasawa, S.~Noda, K.~Kojima, R.~Koyama, F.~Sugai, S.~Nozawa, Y.~Kakiuchi,
  K.~Okada, and M.~Inaba, ``Achievement of dynamic tennis swing motion by
  offline motion planning and online trajectory modification based on
  optimization with a humanoid robot,'' in \emph{Proceedings of the 2016
  IEEE-RAS International Conference on Humanoid Robots}, 2016, pp. 1094--1100.

\end{thebibliography}
}

\end{document}